\documentclass{article}
\usepackage{booktabs} 
\usepackage{times}  
\usepackage{helvet}  
\usepackage{courier}  
\usepackage{url}  
\usepackage{graphicx}  
\usepackage{subfigure}
\usepackage{bm}
\usepackage{amssymb}
\usepackage{array}
\usepackage{amstext}
\usepackage{amsfonts}
\usepackage{cite}
\usepackage{amsmath,amssymb,amsfonts}
\usepackage{algorithmic}
\usepackage{graphicx}
\usepackage{textcomp}
\usepackage{xcolor}
\usepackage{times}
\usepackage{soul}
\usepackage[utf8]{inputenc}
\usepackage[small]{caption}
\usepackage{times}
\usepackage{booktabs} 

\newcommand{\Matrix}[1]{\bm{\mathbf{#1}}}
\usepackage[ruled,vlined]{algorithm2e}
\DeclareMathOperator*{\argmin}{arg\,min}

\def\BibTeX{{\rm B\kern-.05em{\sc i\kern-.025em b}\kern-.08em
    T\kern-.1667em\lower.7ex\hbox{E}\kern-.125emX}}
\usepackage{comment}

\begin{document}
\title{Visual-Texual Emotion Analysis with \\ Deep Coupled Video and Danmu Neural Networks}


\author{Chenchen Li$^{1}$ , Jialin Wang$^{2}$, Hongwei Wang$^{1}$, Miao Zhao$^{2}$, \\
    Wenjie Li$^{2}$, Xiaotie Deng$^3$ \\
$^1$Department of Computer Science, Shanghai Jiao Tong University\\
$^2$Department of Computing, The Hong Kong Polytechnic University\\
$^3$School of Electronics Engineering and Computer Science, Peking University\\
lcc1992@sjtu.edu.cn, wangjialin@hust.edu.cn, \\wanghongwei55@gmail.com, mzhao.ny@gmail.com, \\ cswjli@comp.polyu.edu.hk, xiaotie@pku.edu.cn\\}

\maketitle

\begin{abstract}
    User emotion analysis toward videos is to automatically recognize the general emotional status of viewers from the multimedia content embedded in the online video stream.
    Existing works fall in two categories:
    1) \textit{visual-based} methods, which focus on visual content and extract a specific set of features of videos.
    However, it is generally hard to learn a mapping function from low-level video pixels to high-level emotion space due to great intra-class variance.
    2) \textit{textual-based} methods, which focus on the investigation of user-generated comments associated with videos. The learned word representations by traditional linguistic approaches typically lack emotion information and the global comments usually reflect viewers' high-level understandings rather than instantaneous emotions.
    To address these limitations, in this paper, we propose to jointly utilize video content and user-generated texts simultaneously for emotion analysis.
    In particular, we introduce exploiting a new type of user-generated texts, i.e., ``danmu'', which are real-time comments floating on the video and contain rich information to convey viewers' emotional opinions.
    To enhance the emotion discriminativeness of words in textual feature extraction, we propose \textit{Emotional Word Embedding} (EWE) to learn text representations by jointly considering their semantics and emotions.
    Afterwards, we propose a novel visual-textual emotion analysis model with \textit{Deep Coupled Video and Danmu Neural networks} (DCVDN), in which visual and textual features are synchronously extracted and fused to form a comprehensive representation by deep-canonically-correlated-autoencoder-based multi-view learning.
    Through extensive experiments on a self-crawled real-world video-danmu dataset, we prove that DCVDN significantly outperforms the state-of-the-art baselines.
\end{abstract}



\section{Introduction}
	In some online video platforms, such as Bilibili\footnote{https://www.bilibili.com} and Youku\footnote{http://www.youku.com}, overlaying moving subtitles on video playback streams have become a featured function on the websites, through which users can share feelings and express attitudes towards the content of videos when they are watching.
	Given an online video clip as well as its associated textual comments, visual-textual emotion analysis is to automatically recognize the general emotional status of viewers towards the video with the help of visual information and embedded comments.
	A precise visual-textual emotion analytical method will promote in-depth understanding on viewers' experience, and benefit a broad range of applications such as opinion mining \cite{Yuan:2013:SIS:2502069.2502079}, affective computing \cite{7259380}, and trailer production \cite{Liu:2016:MGM:2964284.2967187}.

	Existing methods for emotion analysis of online videos can be divided into two categories according to the types of input data.
	The first class of methods is \textit{visual-based}, i.e., they take the visual content in videos as input, and perform emotion analysis based on the visual information.
	Typically, in visual-based methods, a specific set of low-level features are extracted from video frames to reveal its underlying emotion \cite{Siersdorfer:2010:APS:1873951.1874060,Borth:2013:SLO:2502081.2502268,You:2015:RIS:2887007.2887061,xCampos}.
	However, visual-based methods exhibit the following limitations:
	1) It is generally hard to learn a mapping function solely from low-level video/image pixels to high-level emotion space due to the great intra-class variance \cite{You:2015:RIS:2887007.2887061,Wang2016BOR}.
	2) It is only feasible to directly apply visual-based methods to images and short videos, as the features of video would increase explosively with its length. Otherwise, visual features need to be periodically sampled.
	3) In visual-based methods, the well-selected visual features are more relevant to the emotion of the video content than the emotion of viewers, which could inevitably dampen their performance in user emotion analysis scenarios.

	As opposed to visual-based methods, the second class of methods is \textit{textual-based}, which utilize user-generated textual comments as input, and extract linguistic or semantic information as features for emotion analysis \cite{7277066,You:2016:CCR:2835776.2835779}.
	Based on their methodologies, existing textual-based methods can further be classified into lexicon-based methods and embedding-based methods.
	Traditional lexicon-based approaches lack considering the syntactic and semantic information of words, hence unable to achieve satisfactory performance in practice.
	Recently, word2vec \cite{NIPS2013_5021}, as a typical example of embedding-based methods, provides an effective way of modeling semantic context of words.
	However, word2vec can only extract semantic proximity of words from texts, while the contextual emotional information is ignored.
	As a result, words with different emotions, such as happy and anger, are mapped to close vectors \cite{tang2014learning}.
	Moreover, it is worth noticing that, most of textual-based methods are based on the global comments for videos (comments that are attached to the videos below), which, unfortunately, can only reflect viewers' high-level understandings on the content rather than their emotion development towards the video.

\begin{figure}
        \centering
        \includegraphics[width=\linewidth]{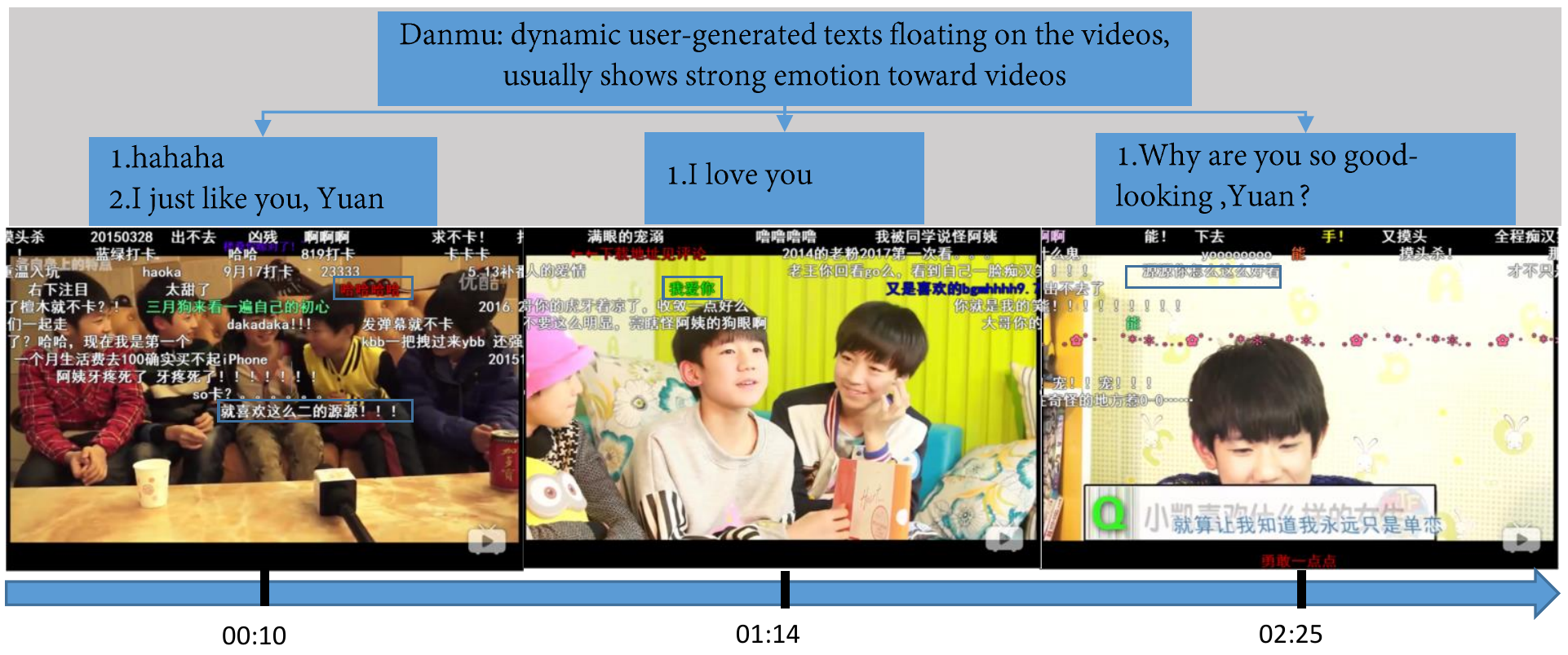}
        \caption{Illustration of danmus associating with a video clip on Bilibili website.}
        \label{fig:danmu_example}
    \end{figure}

	To address aforementioned limitations, in this paper, we consider analyzing viewer's emotion towards online videos by utilizing a new types of textual data, known as ``danmu''.
	Unlike the traditional global comments gathered in a comment section below the videos, danmu is the real-time comments floating on the video in the snapshot, moving along with video playback.
	Viewers can watch the video while sending comments and reading other viewers' comments simultaneously.
	An example of danmu screenshot is illustrated in Figure \ref{fig:danmu_example}.
	Generally, as viewers can express their emotion without any delay, danmus are real-time commentary subtitles and play an important role in conveying emotional opinion from the commentator to other viewers.
	Compared with global comments, danmus have two distinguished characteristics:
	1) Danmus are highly correlated with the specific moments in video clips.
	2) Danmus are generally not distributed uniformly over the whole video.
	In fact, the distribution pattern of danmus reflects the development of the viewers' emotion, e.g., emotion burst, which could greatly facilitate emotion analysis tasks.

  Given danmu as the new source of data, we propose a novel visual-textual emotion analysis model, named \textit{Deep Coupled Video and Danmu Neural networks} (DCVDN).
	DCVDN takes both video frames and associated danmus as input data and aims to construct a joint emotion-oriented representation for viewers' emotion analysis.
	Specifically, for each video clip, we first perform clustering on all of its danmus according to their burst pattern.
	Each set of clustered danmus are aggregated into one danmu document as nearby danmus express viewers' attitudes towards similar video content at a specific moment.
	In DCVDN, to overcome the limitation of emotion-unaware textual-based methods, we propose a novel textual representation learning method, called \textit{Emotional Word Embedding} (EWE), to learn textual features from danmu documents.
	The key idea of EWE is to encode emotional information along with the semantics into each word for joint word representation learning, which is proved to be able to effectively preserve the original emotion information in texts during learning process.
	In addition, we also extract video features from video frames synchronized with the burst points of danmu.
	As viewer's emotion can be reflected as a joint expression of both video content and danmu texts, in this work, we intend to explore the learning of highly non-linear relationships that exist among the visual and textual features. In DCVDN, a joint emotion-oriented representation is developed over the space of video and danmu, by utilizing a Deep Canonically Correlated Auto-Encoder (DCCAE) to achieve multi-view learning for emotion analysis.

  It's also noticeable that {\bfseries each video only has one lable in this work}.
  We know that the emotion of one video may be a mixture of serveral ones with different levels, thus the output of our propose model is the probability over the seven classes
   of the emotions.
  However, our goal is to predict the main emotion of each video.

	To evaluate our proposed DCVDN, we collect video clips and their associated danmus from Bilibili, one of the most popular online video websites in China.
	Our video-danmu dataset consists of 4,056 video clips and 371,177 danmus, in which each example is associated with one of seven emotion classes: happy, love, anger, sad, fear, disgust, and surprise.
	We compare our DCVDN with 14 state-of-the-art baselines by conducting extensive experiments on the video-danmu dataset, and the results demonstrate that DCVDN achieves substantial gains over other visual-based or textual-based methods.
	Specifically, DCVDN outperforms visual-based baselines by $7.03\%$ to $241.59\%$ on \textit{Accuracy} and by $2.52\%$ to $326.99\%$ on \textit{Precision}, and outperforms textual-based baselines by $37.41\%$ to $78.39\%$ on \textit{Accuracy} and by $27.64\%$ to $87.63\%$ on \textit{Precision}.


	\begin{figure*}
 		\includegraphics[width=\textwidth]{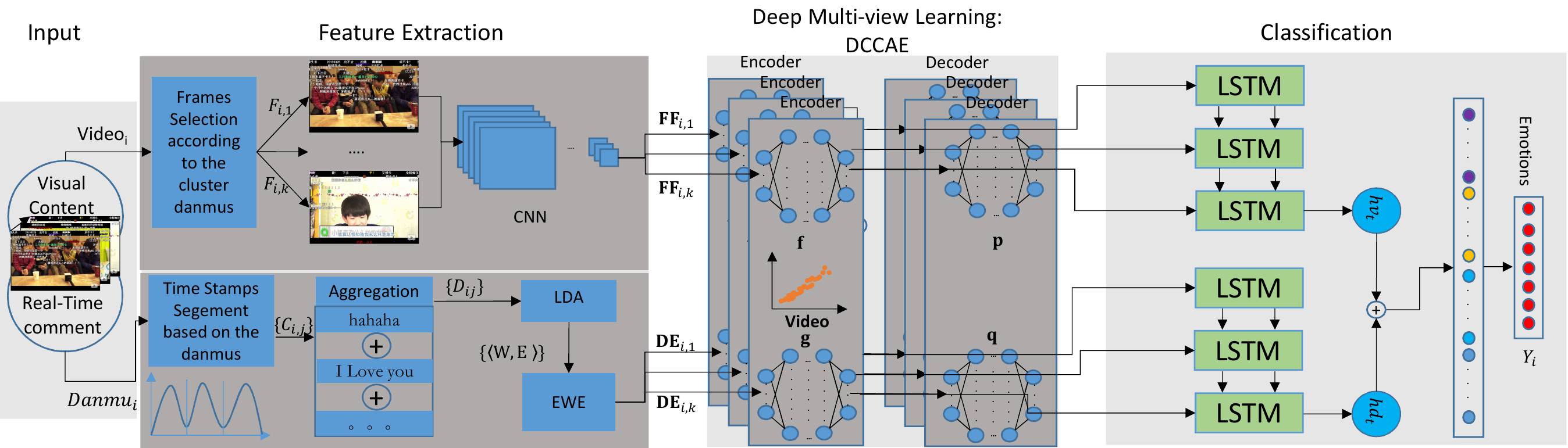}
		\caption{The framework of DCVDN: the video and associated danmus are clustered based on danmus' burst pattern, video segments and danmu documents are synchronized in time, visual and textual features are extracted respectively by CNN and EWE, and finally joint representations are learned by DCCAE for classification. }
		\label{fig:framework}
	\end{figure*}

\section{Related Work}
	Among textual-based methods for emotion analysis, lexicon \cite{7277066,Hu:2013:USA:2488388.2488442} has been widely used due to its simplicity. However, lexicon-based methods cannot exploit the relationship between words. Recently, distributed representations of words have emerged and successfully proliferated in language models and NLP tasks \cite{NIPS2013_5021,You:2016:CCR:2835776.2835779}, which can encode both syntactic and semantic information of words into low-dimensional vectors to quantify and categorize semantic similarities between words.
  Most word embedding models typically represent each word using a single vector, making them indiscriminative under different emotion circumstances. Aware of this limitation, some multi-prototype vector space models have been proposed \cite{Liu:2015:TWE:2886521.2886657,wang2017joint,Reisinger:2010:MVM:1857999.1858012}.
  \cite{Liu:2015:TWE:2886521.2886657} uses latent topic models to discriminate word representations by jointly considering words and their contexts.
  \cite{Reisinger:2010:MVM:1857999.1858012} uses a mix of unsupervised and supervised techniques to learn word vectors capturing semantic term–document information as well as rich sentiment content.
  Distinguishable from existing works, our EWE first uses Latent Dirichlet Allocation (LDA) \cite{Blei:2003:LDA:944919.944937} to infer emotion labels and then incorporates them along with word context in representation learning to differentiate words under different emotional and semantic context.
  There are also various topic models on sentiment analysis \cite{Lin:2009:JSM:1645953.1646003,Li:2010:SAG:2898607.2898826}.
  \cite{Lin:2009:JSM:1645953.1646003} proposes a novel probabilistic modeling framework based on LDA to detect sentiment and topic simultaneously from text.
  \cite{Li:2010:SAG:2898607.2898826} observes that sentiments are dependent on local context, and relaxes  the  sentiment independent assumption.
  It considers the sentiments words as a Markov chain.

	There are also quite a lot of works on visual sentiment analysis. For example, \cite{Siersdorfer:2010:APS:1873951.1874060,7277066} use low-level image properties, including pixel-level color histogram and Scale-Invariant Feature Transform (SIFT), as the features to predict the emotion of images.
  \cite{Borth:2013:SLO:2502081.2502268,Yuan:2013:SIS:2502069.2502079} employ middle-level features, such as visual entities and attributes, as the features for emotion analysis.
  \cite{You:2015:RIS:2887007.2887061,You:2016:BLS:3015812.3015857} utilize Convolutional Neural Networks (CNNs) to extract high-level features through a series of nonlinear transform, which have been proved surpassing other models with low-level and mid-level features \cite{You:2016:BLS:3015812.3015857}.
  \cite{you2017visual} think that the local areas are pretty relevant to human's emotional response to the whole image and proposed model to utilize the recent studies attention mechanism to jointly discover relevant local regions and build a sentiment classifier on top of these local regions.
  \cite{NIPS2013_5204} presented a new deep visual-semantic embedding model trained to identify visual objects using both labeled image data as well as semantic information gleaned from the unannotated text.

	To combine visual and textual information, recent years have witnessed some preliminary effort on multimodal models. For example, \cite{You:2016:CCR:2835776.2835779,7277066} employ both text and images for sentiment analysis. \cite{7277066} employs Deep Boltzmann Machine (DBM) to fuse features from audio-visual and textual modalities, while \cite{You:2016:CCR:2835776.2835779} employs cross-modality consistent regression. Moreover, Deep Neural Network (DNN) based approaches \cite{7277066} are generally used for multi-view representation fusion.
	Prior works have shown the benefits of multi-view method on emotion analysis \cite{You:2016:CCR:2835776.2835779}.
  One step advanced in our work, we employ DCCAE \cite{Wang:2015:DMR:3045118.3045234}, which combines autoencoder and canonical correlation to obtain unsupervised representation learning by jointly optimizing the reconstruction errors minus canonical correlation between extracted features in multiple views.
	Autoencoder is a useful tool for the representation learning, in which the objective is to learn a compact representation that best reconstructs the inputs \cite{ngiam2011multi} via unsupervised learning.
  \cite{kiros2014unifying} introduced an encoder-decoder pipeline that learns (a): a multimodal joint embedding space with images and text and (b): a novel language model for decoding distributed representations from our space.
  \cite{rasiwasia2010new}
  Canonical correlation analysis (CCA) \cite{hotelling1936relations} can maximize the mutual information between different modalities and has been justied in many previous works \cite{hardoon2004canonical,rasiwasia2010new,socher2010connecting}.
  \cite{rasiwasia2010new} makes (CCA) to learn the correlations between visual features and textual features for image retrieval.
  \cite{socher2010connecting} uses a variant CCA to learn a mapping between textual words and visual words.
  Although multi-view methods have been studied extensively, there only exists few works on emotional analysis \cite{You:2016:CCR:2835776.2835779,Wang:2014:MSA:2632856.2632912,Cao:2016:CPS:2962943.2962993}.
  For example, \cite{Wang:2014:MSA:2632856.2632912} proposed a novel Cross-media Bag-of-words Model (CBM) for Microblog sentiment analysis. It represented the text and image of a Weibo tweet as a unified Bag-of-words representation.
\section{Visual-Textual Emotion Analysis}
	In this section, we discuss proposed DCVDN with details.
	We first provide model overview and then introduce video and danmu preprocessing, EWE, DCCAE and classification in the subsequent subsections, respectively.
  Each video is with one label, thus we are solving a one-label classification problem.

\subsection{DCVDN Overview}
		Figure \ref{fig:framework} depicts the framework of DCVDN, which consists of three modules: preprocessing and feature extraction, multi-view representations learning, and classification.

		The first module is committed to preprocess the inputs and extract visual and textual features. It is observed that, for each video clip, danmus are likely to burst around some key video frames.
		The distributions of danmus usually reflect the emotion evolution of the viewers and nearby danmus are more likely to express emotions towards the same video content.
		Therefore, for each video clip, we cluster all of its associated danmus according to their burst pattern.
		Utilizing the results, we aggregate the danmus in each cluster into one document, since it is more effective to analyze longer document rather than shorter ones, which are typically semantic and emotion-ambiguous.
        Afterwards, we aim to learn emotion-aware word embedding and document embedding for each word and each danmu document, respectively.
		Correspondingly, we propose EWE, which combines semantic and emotional information of each word to enhance emotion discriminativeness.
		For videos, we synchronize the selection of the frames corresponding to the burst points of danmus and focus on feature extraction from those selected frames, which are more important and relevant than others to invoke viewers' emotion burst.
		We apply pre-trained Convolutional Neural Networks (CNN) (e.g., VGGNet \cite{simonyan2014very}) to extract features of the video frames as CNN has been proved to achieve the state-of-the-art performance on sentiment analysis \cite{You:2016:BLS:3015812.3015857}.
		These danmu document embeddings and CNN features will be fed into following DCCAE for further joint representation learning.

		The second module is the multi-view representation learning for information fusion between video and danmu. The documents of danmu have highly direct correlations with viewers' emotion and video frames can provide robust background information with appropriate guidance. In DCVDN, for each pair of danmu document and corresponding video frame, we employ DCCAE to learn a multi-view representation in an unsupervised way. A set of obtained multi-view representations will be fed into the following classification module as the input features. From implementation point of view, unsupervised joint representation learning ahead of supervised classification helps avoid complicated end-to-end model training, which effectively facilitate the convergence of training process in practice.

		The last module refers to the classification task.
		It is clear that for each video clip, the multi-view representations output from the second module are still in time series, each corresponding to a clustered time period in the video.
		Hence, Long Short-Term Memory (LSTM) is adopted to address the time dependency across those features. The output of LSTM is treated as the ultimate emotion-aware embedding for each video clip, and eventually fed into softmax function to obtain the target emotion prediction.

	\subsection{Preprocessing and Feature Extraction}
		In this subsection, we discuss the preprocessing and feature extraction methods for danmus and videos in detail. The whole process is also shown in Algorithm \ref{alg:feature_extraction}.

              \setlength{\textfloatsep}{6pt}
		\begin{algorithm}[t]
    			\caption{Preprocessing and Feature Extraction}
    			\KwIn{Videos = \{ $video_1$, $\ldots$, $video_N$ \}, Danmus =  \{ $danmu_1$, $\ldots$, $danmu_N$\}. }
    			\KwOut{$\bm{\mathbf{DEs}}$, $\bm{\mathbf{FFs}}$.}
    			\For{$i \leftarrow 1$ \KwTo $N$}{
        			$C_{i,1}$, $\ldots$ ,$C_{i, K}$ = $\operatorname{K-means}(danmu_i)$; \\
        			\For{$j \leftarrow 1$ \KwTo $K$}{
            				$D_{i,j}$ = $\operatorname{Aggregate}(C_{i,j})$; \\
        			}
    			}
    			$\langle \bm{\mathbf{W}}$, $\bm{\mathbf{E}} \rangle$ = $\operatorname{LDA}(\{ D_{i, j} \}_{1 \leq i \leq N, 1 \leq j \leq K} )$\\
    			$\{ \bm{\mathbf{DE_{i,j}} } \}_{1 \leq i \leq N, 1 \leq j \leq K}$ =
    			$\operatorname{EWE}$($\langle \bm{\mathbf{W}}$, $\bm{\mathbf{E}} \rangle$)\\
    			\For{$i \leftarrow 1$ \KwTo $N$}{
        			\For{$j \leftarrow 1$ \KwTo $K$}{
            				$F_{i, j}$ = $\operatorname{FrameSelect}(video_i, C_{i,j})$; \\
            				$\bm{\mathbf{FF_{i,j}}}$ = $\operatorname{CNN}(F_{i,j})$;
        			}
    			}
    			$\bm{\mathbf{DEs}}$ = $ \{ \bm{\mathbf{DE}}_{1,1}; \ldots; \bm{\mathbf{DE}}_{N,K} \} $;\\
    			$\bm{\mathbf{FFs}}$ = $ \{ \bm{\mathbf{FF}}_{1,1}; \ldots; \bm{\mathbf{FF}}_{N,K} \} $; \\
    			\label{alg:feature_extraction}
		\end{algorithm}

\subsubsection{Preprocessing and Feature Extraction on Danmu}
			As aforementioned, danmu is a kind of timely user-generated comment with non-uniform distribution over entire video. The distribution of danmus reflects user engagement to the video content and the video content at burst points of danmus is typically more attractive to viewers than other parts.
			Aware of this phenomenon, we apply K-means algorithm to segment danmus into a set of clusters according to their burst pattern and aggregate all danmus in the same cluster into a danmu document.
			Formally speaking, consider a dataset with a total of $N$ videos, denoted by $V = \{video_1, \ldots, video_N\}$.
			Each $video_i$ is associated with a collection of danmus, denoted by $danmu_i = \{ (s_{i,1}, \textit{offset}_{i,1}), \ldots, (s_{i, n_i}, \textit{offset}_{i, n_i}) \}$,
 where $s_{i,j}$ represents the text of $j$-th danmu for $video_i$, ${\textit{offset}_{i,j}}$ represents the emergence moment of $s_{i,j}$ relative to the beginning of $video_i$ ($\textit{offset}_{i,j} < \textit{offset}_{i,h}$, if $j < h$), and $n_i$ is the total number of danmus in $video_i$.
			For each $video_i$, we aim to find a $K$-partition \{$C_{i,1}$, \ldots, $C_{i,K}$\} satisfying
			\begin{equation}
    				\argmin \limits_{C_{i,1}, \ldots, C_{i,K}} \sum \limits_{j=1}^{K} \sum \limits_{h=1}^{|C_{i,j}|}\left | \textit{offset}_{i,\sum_{t=1}^{j-1}{|C_{i,t}|}+h} - \mu_{i,j}\right|^{2},
			\end{equation}
			where $\cap_{1 \leq j  \leq K} C_{i,j} = \emptyset$ and for each cluster $j$, we have
			$$\mu_{i, j} = \frac{\sum_{h=1}^{|C_{i,j}|}\textit{offset}_{i,\sum_{t=1}^{j-1}{|C_{i,t}|}+h}}{|C_{i,j}|}$$,
			which is the centroid of cluster $j$ and is also treated as the burst point of cluster $C_{i,j}$.
			Once clusters are formed, we obtain the danmu document set for $video_i$ by aggregating all danmus in same clusters, i.e., $D_{i}$ = \{$D_{i,1}$, \ldots, $D_{i,K}$\}, where each $D_{i,j}$ corresponds to a danmu document, i.e., $D_{i,j} = \oplus_{1 \leq h \leq |C_{i,j}|} s_{i,\sum_{t=1}^{j-1}{|C_{i,t}|}+h}$, and the document set includes all danmu documents associated with $video_i$, i.e., $D_{i}$ = $\cup_{1 \leq j \leq K} D_{i,j}$.

			To extract the textual features from danmu to enhance emotion discriminativeness, we correspondingly propose emotion-based Latent Dirichlet Allocation (eLDA) \cite{Blei:2003:LDA:944919.944937} and EWE to first learn the emotional embedding of each word and then derive the emotional document embedding for each danmu document $D_{i,j}$. We will discuss the details in the next subsection.

		\subsubsection{Preprocessing and Feature Extraction on Video}
			For videos, we exploit the clustering information in danmus to select frames for visual feature extraction.
			Specifically, we draw out the video frames corresponding to the burst points of danmu clusters in each video clip as they are more attractive to the viewers. In this way, the video frames and danmu documents are synchronized in time. Formally, for each video $video_i$, based on danmu cluster partition $\textit{danmu}_i$ =  $\{C_{i,1}$, \ldots, $C_{i,K}\}$, we select the key frame $F_{i,j}$ at the time moment of burst point $\mu_{i,j}$ to represent the basic visual content of cluster $j$.
			As the result, we would get a set of frames for $video_i$, i.e., $\{F_{i, 1}$, \ldots, $F_{i, K}\}$, in one-to-one correspondence to the danmu clusters $\{C_{i,1}, \ldots, C_{i, K}\} $.

			Previous work \cite{You:2016:BLS:3015812.3015857} has shown that visual features extracted by CNN networks can achieve satisfactory performance for emotion analysis. Therefore, in this work, we employ the pre-trained CNN, i.e., VGG-Net fc-7 \cite{simonyan2014very}, for visual feature extraction from each video frame $F_{i, j}$. Basically, danmu texts could explicitly deliver viewers' opinions and video frames would provide the supportive background information of emotion-relevant content.

	\subsection{Danmu Document Embedding Learning}
		In this subsection, we discuss the emotion-oriented embedding learning for word and danmu documents.
		We first introduce eLDA method to estimate the emotion label of each word, and then discuss the details of proposed EWE model, which aims to combine emotion and semantic information to learn a comprehensive and effective word embedding to facilitate viewers' emotion analysis.

		\subsubsection{eLDA}
			LDA \cite{Blei:2003:LDA:944919.944937} is an unsupervised model and is commonly used to infer the topic label for words and documents. Inspired by LDA for topic analysis, in this work, we exploit it to infer the emotion labels by considering danmu documents as random mixture over latent emotions and each emotion is characterized by a distribution over words. Particularly, each danmu document \text{$D_{i,j}$} is represented as a multinomial distribution $\bm{\mathbf{\Theta}}_D$ from Dirichlet distribution $\operatorname{Dir}(\bm{\mathbf{\alpha}})$ over a set of emotions, each emotion is usually represented as a multinomial distribution $\bm{\mathbf{\Pi}}_{l}$ over a set of vocabulary from $\operatorname{Dir}(\bm{\mathbf{\beta}})$.
			The generative process is defined formally as follow:
			\begin{itemize}
    				\item For each danmu document $D_{i,j}$, choose a multinomial distribution $\bm{\mathbf{\Theta}}_D$ over the emotions from $\operatorname{Dir}(\bm{\mathbf{\alpha}})$;
    				\item For each emotion $l$, choose a multinomial distribution $\bm{\mathbf{\Pi}}_{l}$ over the words from $\operatorname{Dir}(\bm{\mathbf{\beta}})$;
    				\item For each word position $t$ in document $D_{i,j}$,
    				\begin{itemize}
        				\item Choose an emotion $l_{t}$ from $\operatorname{Multinomial}(\bm{\mathbf{\Theta}}_D)$;
        				\item Choose a word $w_t$ from the  $\operatorname{Multinomial}(\Matrix{\Pi}_{l_t})$.
    				\end{itemize}
			\end{itemize}

			From implementation perspective, in order to effectively infer emotions, we need prior knowledge of the emotional ground truths of some words.
			When determining the emotion of a word, if the word exists in our emotion lexicon, we choose to use its corresponding emotion in lexicon, otherwise, we choose the emotion according to the probabilities of $\operatorname{Multinomial}(\mathbf{\Theta}_{D})$.
			Considering that danmu culture (sometimes called manga and anime) is mainly popular among youngsters, the authentic word emotions are somewhat different from the common sense in the existing lexicon. Therefore, it is desired to build a new lexicon specifically for the manga and anime culture.
			We spend great effort to construct such kind of lexicon, which consists of $1,592$ network-popular words and $1,670$ emoticons. Emoticons are the textual portrayals of a user's moods or facial expressions in the form of icons. For example, $^\land{_{-}}^{\land}$  represents happiness and $\top \_ \top$ stands for crying.
			We select these focused words and emoticons according to their occurrence frequency in our dataset.

            It is worth pointing out that we cluster the emotion distribution into certain number of classes and treat the result of clustering as the final emotion label for each word, rather than directly use the emotion with the maximal probability as adopted in TWE \cite{Liu:2015:TWE:2886521.2886657}. Specifically, suppose we obtain the emotion distribution $\mathbf{\Theta}_{i, j}$ of each word $w_{i,j}$ (the $j$-th word in the damu aggregation of $video_{i}$) after the interference of eLDA model. Then we use K-means algorithm to cluster these emotion distributions, which aims to find a $KE$-partition $ \{ CE_1, \ldots, CE_{KE} \}$ satisfying
            \begin{equation}
                    \argmin \limits_{CE_{1}, \ldots, CE_{KE}} \sum \limits_{k=1}^{KE} \sum \limits_{\mathbf{\Theta}_{i, j} \in CE_{k}} \left | \mathbf{\Theta}_{i, j} - \mathbf{\eta}_{k} \right|^{2}
            \end{equation}
            where $\cap_{1 \leq k \leq KE} C_{k} = \emptyset$ and  $\eta_{k}$ is the centroid of cluster $k$. The new emotion label $l_{i,j}$ of $w_{i,j}$ is $k$ if $\mathbf{\Theta}_{i, j} \in CE_{k} $. The reason for the clustering is that the number of emotion labels is generally small (7 for emotion classification tasks) and we can't fully explore the information hidden in the distributions with such a few labels. To avoid the dilemma, we recluster the distributions into more classes in order to make  the new labels more discriminative. The new labels would be used to learn EWE at a later time.


		\subsubsection{Emotional Word and Document Embeddings}
			\begin{figure}
    				\centering
    				\includegraphics[width=0.8\textwidth]{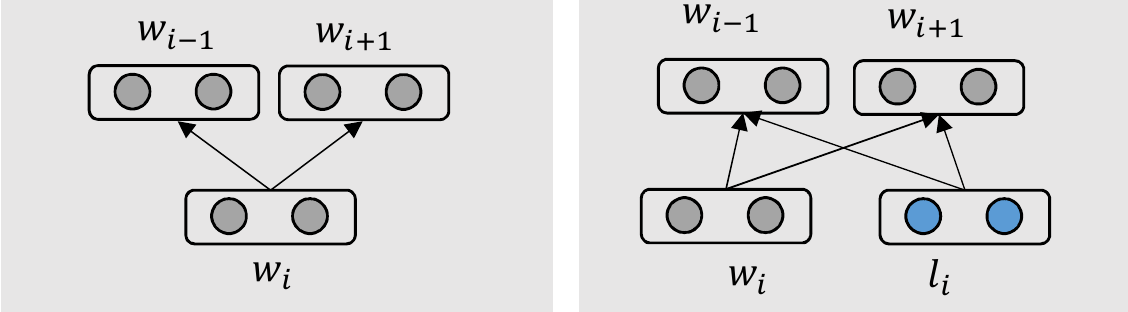}
    				\caption{Skip-Gram and EWE models. Gray and blue circles represent word and emotion embeddings, respectively.}
    				\label{fig:jetewe}
			\end{figure}

			Word embedding, which represents each word using a vector, is widely used to capture semantic information of words. Skip-Gram model \cite{NIPS2013_5021} is a well-known framework for word embedding, which learns word representation that is useful for predicting context words in a sliding window when given the target word. Formally, given a word sequence $\{w_1, \ldots, w_{T}\}$, the objective of Skip-Gram is to maximize the average log probability
			\begin{equation}
        			\mathcal{L}(D) = \frac{1}{T} \sum_{t=1}^{T} \sum_{-k \leq c \leq k, c \neq 0} \log p(w_{t+c}|w_t),
    				\label{equ:skipgram:loss}
			\end{equation}
			where $k$ is the context window size for the target word, which can be a function of the centered word $w_{t}$.
			Probability $p( w_c | w_t)$ is defined as a softmax function as follows:
			\begin{equation}
    				P(w_c|w_t) = \frac{\exp{(\bm{\mathbf{w}}_c^{\top}\bm{\mathbf{w}}_t )}}{\sum_{\bm{\mathbf{w}} \in W} \exp{(\bm{\mathbf{w}}^{\top}\bm{\mathbf{w}}_t )}},
    				\label{equ:skipgram:softmax}
			\end{equation}
			where $\bm{\mathbf{w}}_t$ and $\bm{\mathbf{w}}_c$ are word vectors of target word $w_t$ and context word $w_c$, respectively, and $W$ is word vocabulary.

                         It is noticed that Skip-Gram model for word embedding focuses on the semantic context and assumes that each word always preserves a single vector, which sometimes is indiscriminate under different emotion circumstances. These facts motivate us to propose a joint emotion and semantics learning model, named EWE. The basic idea of EWE is to preserve emotion information of words when measuring the interaction between target word $w_{t}$ and context word $w_{c}$. In this way, a word with different associated emotions would correspond to different embeddings so as to effectively enhance the emotion discriminativeness of each word.

			Specifically, rather than solely using the target word $w_t$ to predict
context words in Skip-Gram, inspired by \cite{Liu:2015:TWE:2886521.2886657}, EWE jointly utilizes $l_{t}$'s as well, i.e., the emotions of the words in the danmu documents.
			EWE aims to learn the representations for words and emotions separately and simultaneously. In particular, it regards each emotion as a pseudo word and consider the occurrence of this pseudo word in all the positions wherever the positioned words are assigned with this emotion. EWE uses both the target word $w_{t}$ and its associated emotion $l_{t}$ to predict context words, as shown in Figure \ref{fig:jetewe}.
			For each target word with its  emotion $\langle w_t, l_t \rangle$, the objective of EWE is to maximize the following average log probability
			\begin{equation}
			   \begin{aligned}
    				\mathcal{L}(D) = \frac{1}{T} \sum_{t=1}^{T} \sum_{-k \leq c \leq k}^{c \neq 0} \Big( \log p(w_{t+c}|w_t) + \log p(w_{t+c}|l_t) \Big),
    			\end{aligned}
			\end{equation}
      where $P(w_c|l_t)$ is similar with $P(w_c|w_t)$
      \begin{equation}
    				P(w_c|l_t) = \frac{\exp{(\bm{\mathbf{w}}_c^{\top}\bm{\mathbf{l}}_t )}}{\sum_{\bm{\mathbf{w}} \in W} \exp{(\bm{\mathbf{w}}^{\top}\bm{\mathbf{l}}_t )}},
    				\label{equ:skipgramemtoion:softmax}
			\end{equation}
      and $\bm{\mathbf{w}}$ and $\bm{\mathbf{l}}$ are the represention vectors of words and emotions respectively. When we minimize the log loss $\log p(w_{t+c}|l_t)$, we consider that $w_t$ is $l_t$.
      The other learning process of emotional word is the same as the textual word. The process is shown in Algorithm \ref{alg:ewe}.
      \begin{algorithm}
          \caption{EWE}
          \KwIn{$W$= [$\langle w_1, l_1 \rangle$, $\ldots$, $\langle w_{N_{seq}}, l_{N_{seq}} \rangle$], $c$, $m$, $\mu$}
          \KwOut{\\
          $\bm{\mathbf{V}}_w$ = [$\bm{\mathbf{v}}_w(w_1)$, $\ldots$, $\bm{\mathbf{v}}_w(w_{N_w})$],\\
          $\bm{\mathbf{V}}_l$ = [$\bm{\mathbf{v}}_l(l_1)$, $\ldots$, $\bm{\mathbf{v}}_l(l_{N_l})$]}
          Initialize randomly a matrix $\bm{\mathbf{V}}_w \in R^{N_w \times m}$  \\
          Initialize randomly a matrix $\bm{\mathbf{V}}_l \in R^{N_l \times m}$  \\
          \For{$i \leftarrow 1$ \KwTo $n$}{
              $L$ = 0 \\
              \textbf{Forward Propagation}: \\
              \For{$j \leftarrow i - c$ \KwTo $i + c$}{
                  \If{$j < 0$ or $j > n$}{
                      $w_j$ = \#
                  }
                  \If{$j \neq i$}{
                      $L$ = $L$ + $\operatorname{log} P (w_j|w_i)$ +  $\operatorname{log} P (w_j|l_i)$
                  }

              }
              \textbf{Backward Propagation}: \\
              $\bm{\mathbf{v}}_w(w_i)$ = $\bm{\mathbf{v}}_w(w_i)$ - $\mu\frac{\partial L}{\partial \bm{\mathbf{v}}_w(w_i)}$ \\
              $\bm{\mathbf{v}}_l(l_i)$ = $\bm{\mathbf{v}}_l(l_i)$ - $\mu\frac{\partial L}{\partial \bm{\mathbf{v}}_l(l_i)}$
          }
          \label{alg:ewe}
      \end{algorithm}
      $N_{seq}$ is the length of the documents, $N_w$ is the number of vocabulary and $N_l$ is the number of emotions, which are the results of clustering of eLDA.
      $c$ is the size of contextual window, $m$ is the user-defined size of representation and $\mu$ is the learning rate.
      $V_w$ and $V_l$ are the representation vectors of words and emotions respectively.

      Emotional word embedding of word $w$ in emotion $l$ is obtained by concatenating the embeddings of $\mathbf{w}$ and $\mathbf{l}$, i.e., $\bm{\mathbf{w}}^{l} = \bm{\mathbf{w}} \oplus \bm{\mathbf{l}}$, where $ \oplus$ is the concatenation operation and the dimension of $\mathbf{w}^{l}$ is double of $\mathbf{w}$ and $\mathbf{l}$.
			Correspondingly, the document embedding in EWE is to aggregate emotional word embeddings of the words in a danmu document. The document embedding of $D_{i,j}$ is defined as $\bm{\mathbf{d}} = \sum_{w \in D_{i,j}} \operatorname{P}(w|D_{i,j}) \bm{\mathbf{w}}^{l}$, where $\operatorname{P}(w|D_{i,j})$ can be the term frequency-inverse document frequency of word $w$ in $D_{i,j}$.

	\subsection{Deep Multi-View Representation Learning}
		In this subsection, we introduce the multi-view representation learning method in DCVDN, which aims to simultaneously utilize video and danmu information to learn a joint representation based on the extracted visual and textual features. Inspired by canonical correlation analysis (CCA) and reconstruction-based objectives, we employ deep canonically correlated autoencoders to fuse the latent features from video and danmu points of views. In particular, DCCAE consists of two autoencoders and optimizes the combination of canonical correlation between the learned textual and visual representations and the reconstruction errors of the autoencoders. The structure of DCCAE is shown in the module of ``Deep Multi-view Learning: DCCAE'' in Figure \ref{fig:framework} and its optimization objective is as follows
		\begin{equation}
    			\begin{gathered}
        			\min_{\bm{\mathrm{W}}_{\bm{\mathrm{f}}}, \bm{\mathrm{W}}_{\bm{\mathrm{g}}}, \bm{\mathrm{W}}_{\bm{\mathrm{p}}}, \bm{\mathrm{W}}_{\bm{\mathrm{q}}}, \bm{\mathrm{U}}, \bm{\mathrm{V}}}
        - \frac{1}{M} \operatorname{tr}(\bm{\mathrm{U}}^{\top} \bm{\mathrm{f}}(\bm{\mathrm{X}})\bm{\mathrm{g}}(\bm{\mathrm{Y}})^{\top}\bm{\mathrm{V}})\\
        + \frac{\lambda}{M} \sum_{i=1}^{M}(\left \| \bm{\mathrm{x}}_i - \bm{\mathrm{p}}(\bm{\mathrm{f}}(\bm{\mathrm{x}}_i)) \right \| + \left \| \bm{\mathrm{y}}_i - \bm{\mathrm{q}}(\bm{\mathrm{g}}(\bm{\mathrm{y}}_i)) \right \|), \\
        \mathbf{s.t.} \bm{\mathrm{u_i}}^{\top} \bm{\mathrm{f}}(\bm{\mathrm{X}})\bm{\mathrm{f}}(\bm{\mathrm{X}})^{\top} \bm{\mathrm{u_i}} = \bm{\mathrm{v_i}}^{\top} \bm{\mathrm{g}}(\bm{\mathrm{Y}})\bm{\mathrm{g}}(\bm{\mathrm{Y}})^{\top} \bm{\mathrm{v_i}} = M, 1 \leq i \leq L,
        		\label{eq:dccae}
    			\end{gathered}
		\end{equation}
		where $\lambda > 0$ is the trade-off parameter, $M$ is the sample size, $\Matrix{X} = [\Matrix{x}_1, \ldots, \Matrix{x}_{M}$] and $\Matrix{Y} = [\Matrix{y}_1, \ldots, \Matrix{y}_{M}$] are the feature matrices of visual and textual viewpoints, each $\Matrix{x}$ and \Matrix{y} referring to the visual and textual features extracted from a damu document and corresponding video frame, respectively.
		Moreover, $\Matrix{f}$, $\Matrix{g}$, $\Matrix{p}$ and $\Matrix{q}$ denote mapping functions implemented in autoencoders. The encoder-decoder pair ($\Matrix{f}$, $\Matrix{p}$) and ($\Matrix{g}$, $\Matrix{q}$) constitute two autoencoders, each for one of two viewpoints. The corresponding parameters in encoding functions $\Matrix{f}$ and $\Matrix{g}$ and decoding functions $\Matrix{p}$ and $\Matrix{q}$ are denoted by $\Matrix{W_f}$, $\Matrix{W_g}$, $\Matrix{W_p}$, and $\Matrix{W_q}$, respectively.
		$\Matrix{U} = [\Matrix{u}_1, \ldots, \Matrix{u}_L]$ and $\Matrix{V} = [\Matrix{v}_1, \ldots, \Matrix{v}_L]$ are the CCA directions that project the outputs of $\Matrix{f}$ and $\Matrix{g}$, where $L$ is the dimensionality of input features to autoencoders.

		Mathematically, the first term of Eq. (\ref{eq:dccae}) is the objective of CCA, while the second and third terms are the losses of autoencoders, which can be understood as adding autoencoders as regularization terms to CCA objective. The constraint is for CCA to ensure the objective is invariant to the scale of $\Matrix{U}$ and $\Matrix{V}$.
		CCA aims to maximize the mutual information between videos and danmus, while autoencoders aim to minimize the reconstruction errors of two views.
		In this way, DCCAE tries to explore an optimal trade-off between the information captured from the reconstruction of each view and the information captured from learning the mutual information of two views. Therefore, it can achieve better representations.
		The output features of two views are $\bm{\mathrm{U}}^{\top} \bm{\mathrm{f}}(\bm{\mathrm{X}})$ and $\bm{\mathrm{V}}^{\top} \bm{\mathrm{g}}(\bm{\mathrm{Y}})$, respectively, which would be as the inputs to two seperated LSTM for the later classification.

	\subsection{Classification}
  In this module, the DCCAE outputs are utilized to do classification.
  As aforementioned, the output representations from DCCAE are still in time series from two modalities.
  To address time dependency across the representations, we feed the two modalities into two separated LSTMs and get the final outputs of two LSTMs, $h_v$ from the video part and $h_t$ from the text part.
  Then we simply concatenate two parts into one, $h_a$ = $h_v \oplus h_t$.
  The obtained representation $h_a$ would eventually be fed into following fully-connected network with a softmax function to obtain the target emotion prediction.
  The classification network is depicted as the rightmost module in Figure \ref{fig:framework}.

\section{Experiments}
	In this section, we carry out extensive experiments to evaluate the performance of DCVDN.
	We first introduce the datasets used for experiments and then compare our model with other 14 baselines, which address on visual, textual and joint features for emotion analysis and videos classification, respectively.

	\subsection{Datasets}
		\begin{table}[t]
		\centering
    			\caption{The basic statistics of the Video-Danmu dataset.}
    			\begin{tabular}{|c|c|c|c|c|c|c|}
    			\hline
    			\multicolumn{3}{|c|}{Number of Videos} & \multicolumn{2}{c|}{Avg. Length} & \multicolumn{2}{c|}{Length Range} \\
    			\hline
    			\multicolumn{3}{|c|}{4,056} & \multicolumn{2}{c|}{82.89s} & \multicolumn{2}{c|}{1.44s - 514.13s} \\
    			\hline
    			Happy & Love & Anger & Sad & Fear & Hate & Surprise \\
    			\hline
    			620 & 877 & 290 & 631 & 647 & 669 & 322 \\
    			\hline
    			\end{tabular}
    			\label{tab:video_statistic}
		\end{table}

		\begin{table}
		\small
		\centering
    			\caption{Number of words and emoticons of different emotion classes in self-built emotion lexicon.}
          \begin{tabular}{|c|c|c|c|c|c|c|}
    			\hline
			\multicolumn{7}{|c|}{Number of Words}\\
			\hline
    			Happy & Love & Anger & Sad & Fear & Hate & Surprise \\
    			\hline
    			90 & 784 & 25 & 132 & 146 & 348 & 67 \\
    			\hline
    			\multicolumn{1}{|c|}{Total} & 1592 & \multicolumn{3}{c|}{Ave. Occurrence Freq.} & \multicolumn{2}{c|}{1550.39}\\
    			\hline
			\hline
			\multicolumn{7}{|c|}{Number of Emoticons}\\
			\hline
                Anger & Disgust & Fear & Shame & Guilt & Joy & Sadness \\
    			\hline
    			417 & 522 & 41 & 192 & 148 & 265 & 85 \\
    			\hline
    			\multicolumn{1}{|c|}{Total} & 1670 & \multicolumn{3}{c|}{Ave. Occurrence Freq.} & \multicolumn{2}{c|}{7.90} \\
    			\hline
    			\end{tabular}
    			\label{tab:word_emoticon_class}
		\end{table}

        \subsubsection{Datasets For Video-Danmu}
        Considering the lack of existing danmu-related dataset, we put great effort to self construct a new dataset, called Video-Danmu\footnote{We are happy to share this dataset to public after the paper gets published.}.
        This dataset include videos and their associated danmus directly crawled from Bilibili website, which is one of the most popular websites providing danmu services in China.
        There are 4,056 videos in the dataset, which last ranging from 1.44 to 514.13 seconds and average at 82.89 seconds.
        We labelled the videos into 7 emotion classes, i.e., happy, love, anger, sad, fear, disgust and surprise, with the help of a group of student helpers in our university.
        Table \ref{tab:video_statistic} shows the basic statistics of the dataset. The number of videos falling in each emotion category is relatively balanced, ranging from 290 to 669 pieces.
        Table \ref{tab:word_emoticon_class} lists the number of words and emoticons belong to each emotion class in self-built emotion lexicon.
        Emoticon is a kind of text expression, like ( $\land$ $\land$ ) show the happiness, $\top \_ \top$ show the crying. They usually directly express the emotion of viewers.
        The average occurrence frequency of the words in our dataset is about $1,550$, which strongly validates their popularity in practice.
        And the average occurrence of emoticons in the dataset is about 8 times.

         \subsubsection{Datasets For Textual Analysis}
         In order to show that our EWE can be applied to other text-based emotion applications as well, we also use two additional text datasets for comparison.
        \begin{itemize}
            \item Incident reports dataset (ISEAR) \cite{parrott2001emotions}: ISEAR contains $7,000$ incident reports obtained from an international survey on emotion reactions. A number of psychologists and non-psychologists, were asked to report situations in which they had experienced all of 7 major emotions (joy, fear, anger, sadness, disgust, shame, and guilt). 
            \item Multi-Domain Sentiment Dataset \cite{blitzer2007biographies}: This dataset contains four-domain (books, dvd, electronic and kitchen \& housewares) reviews of productions from Amazons. It consists of $8,000$ reviews, including
            $4,000$ positive reviews and $4,000$ negative reviews.
        \end{itemize}

	\subsection{Baselines}
		We compare proposed DCVDN with other 14 baselines, which can be divided into four categories, i.e., visual-based, textual-based, multi-view learning and video classification methods. We also compare proposed EWE with other textual emotion analysis baselines.

Visual-based baselines:
		\begin{itemize}
    			\item GCH/LCH: use low-level features (64-bin global color histogram features (GCH) and 64-bin local color histogram (LCH)) as defined in \cite{Siersdorfer:2010:APS:1873951.1874060}.
    			\item Caffenet: An ImageNet pre-trained AlexNet \cite{Campos:2015:DDS:2813524.2813530} followed by fine tuning.
    			\item PCNN: Progressive CNN \cite{You:2015:RIS:2887007.2887061}.
		\end{itemize}
		Note that in all above approaches, the image features of selected frames will be fed into LSTM for final classification.

		Textual-based baselines:
		\begin{itemize}
    			\item Lexicon method: We count the number of words belonging to each emotional class in each document. Then we choose the emotion class with the largest count as the result.
    			\item eLDA: Aggregate all danmus of a video into one document and infer the emotion distribution of the document. Choose the emotion class with the largest probability as the result.
    			\item Word embedding: Learn word representations by Skip-Gram model \cite{NIPS2013_5021}.
    			\item Topical word embedding (TWE): Learn word representations by TWE model \cite{Liu:2015:TWE:2886521.2886657}, which jointly utilizes the target word and its topic to predict context words.
                \item SSWE \cite{tang2014learning}: Sentiment-Specific word embedding for sentiment classification.
		\end{itemize}

		Multi-view learning baselines:
		\begin{itemize}
    			\item Simple-Con: Concatenate the features from different views.
    			\item DistAE: A joint learning \cite{Wang:2015:DMR:3045118.3045234}, the objective of which is a combination of reconstruction errors of autoencoders of different views and the average discrepancy across the learned projections of multiple views.
        \end{itemize}

        Video classification baselines:
        \begin{itemize}
            \item Conv3D \cite{tran2015learning}: 3D Convolutional neural networks.
            \item Temporal \cite{simonyan2014two}: Optical flows of the frames in the video, widely used for actions recognition.
            \item Temporal + Spatial \cite{simonyan2014two}: Use CNN to extract the spatial features, and average the temporal features and spatial features.
        \end{itemize}

	\subsection{Parameter Settings}
		For the set of danmus in each video, we divide it into $10$ clusters and aggregate each part into one document.
    While the number of clusters is user-defined parameter, and $10$ performans well in our experiments, thus we recommend it.
    In EWE, the dimensions of word vectors and emotion vectors are $128$, therefore the dimensions of emotional word and document embeddings are $256$.
		The visual features are extracted by VGG-Net fc-7, which results in $4,096$ features.
		In multi-view learning module, the two autoencoders in the DCCAE are $3$-layer, in which the size of the middle layer is $256$ and the size of other layers is equal to the inputs.
		In the classification module, We use LayerNorm LSTM \cite{ba2016layer} here. The length of LSTM is set to $10$, forget-bias is $0.1$, and hidden layer size is $2,048$. The following fully-connected network has 2 layers, with the size of the hidden layer as $4,096$. We focus on \textit{Accuracy} and \textit{Precision} as the evaluation metrics in our experiments.

 \subsection{Case Studies}
        In this subsection, we first present one example to show that the key frames are strongly related to the burst points of danmus, then we present three prediction examples to validate the superiority of DCVDN over other baselines.
        \subsubsection{An Example of Key Frames and Danmus}
\begin{figure}
            \centering
            \includegraphics[width=\textwidth]{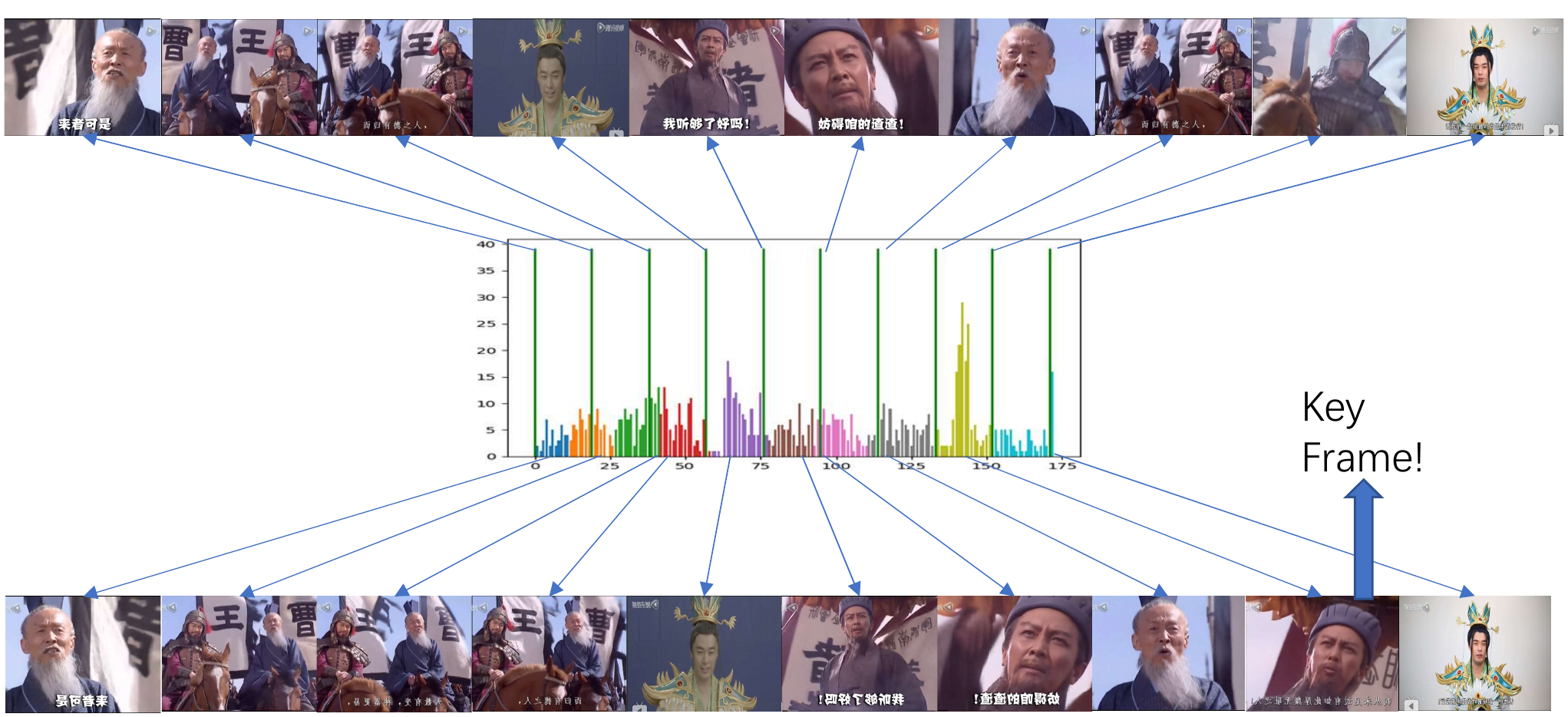}
            \caption{An example of the relation between burst points of danmus and selected key frames.
            The above frame sequence is achieved by even choosing, and the below frame sequence is achieved by our clustering method. The middle chart shows the change in the amount of danmus appearing in each second. }
            \label{fig:bursting}
        \end{figure}

        We provide an example of the relationship between the burst points of danmus and the selected key frames as shown in Fig. \ref{fig:bursting}, in order to show that our clustering approach can select more important frames with the help of danmus.
        The most famous clip of this video is that an ancient minister just abused his opponent. "I've never seen anyone so brazen!" his opponent angrily said and died then.
        The above frame sequence is achieved by even chosen from the video, and the lower frame sequence is achieved by our clustering method based on damu burst pattern. The middle chart shows the change in the amount of danmus appearing in each second. Our method successfully finds the key frame, which more comprehensively reflect the content background.
        It is also evident in the middle chart that the amount of danmus always change over time and the changes are strongly related to the audiences' interest.
        Our clustering method nearly selects the frame with most danmus in each time interval.
        Moreover, our method selects the frame with the most important words, "I've never seen anyone so brazen!".
        And as shown in the chart, the selected frame corresponds to the time moment with most danmus.
        By contrast, the uniform selection misses the frame, which is not always effective in practice.

        \subsubsection{Three Examples of Prediction Results}
        \begin{figure}
        \centering
            \subfigure[A wacky video with common visual content and scaring BGM. The label is ``Disgust''. VGG predicts Love, EWE predicts Fear and DCVDN predicts correctly. ]{\label{fig:subfig:a}
                \includegraphics[width=0.9\textwidth]{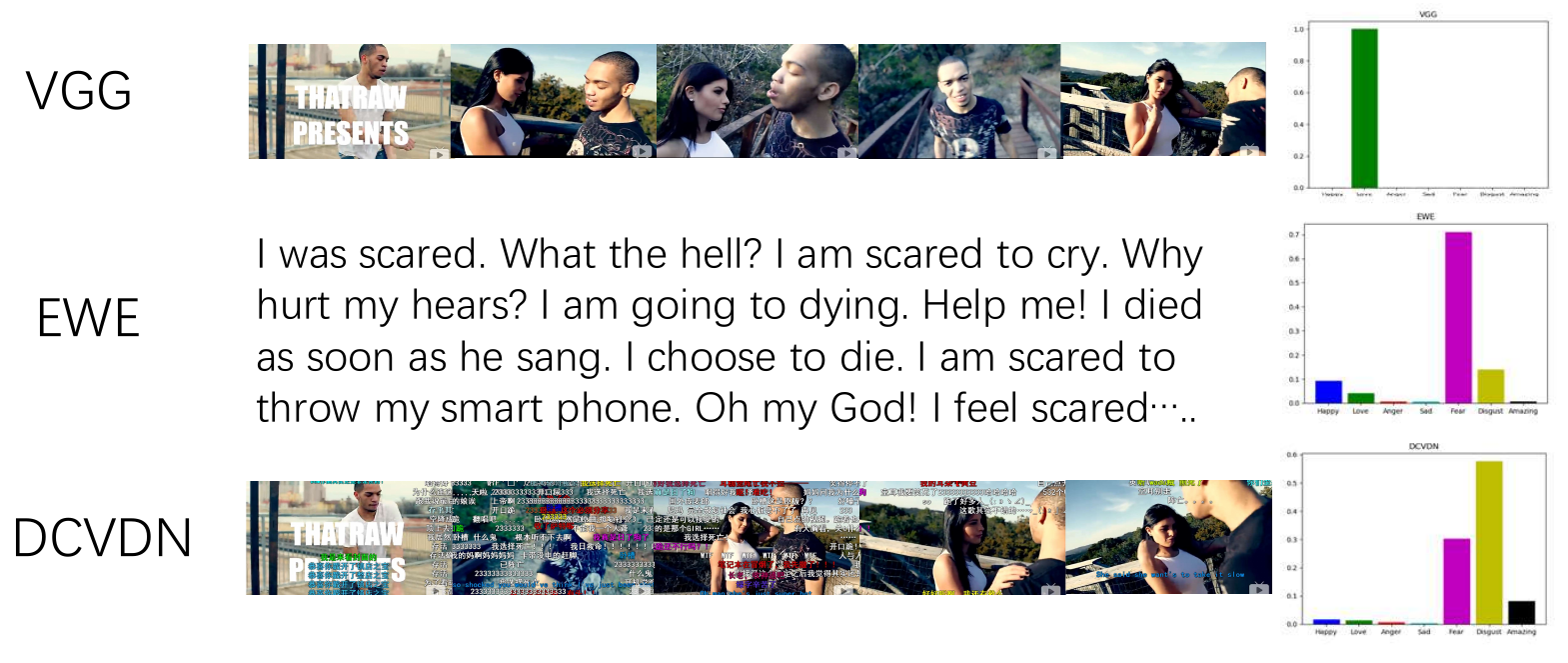}}
            \subfigure[Two audiences made proposes to the tennis stars and got different responses. The label is ``Happy''. VGG and DCVDN predict correctly, but EWE considers it's ``Fear''. ]{\label{fig:subfig:b}
                \includegraphics[width=0.9\textwidth]{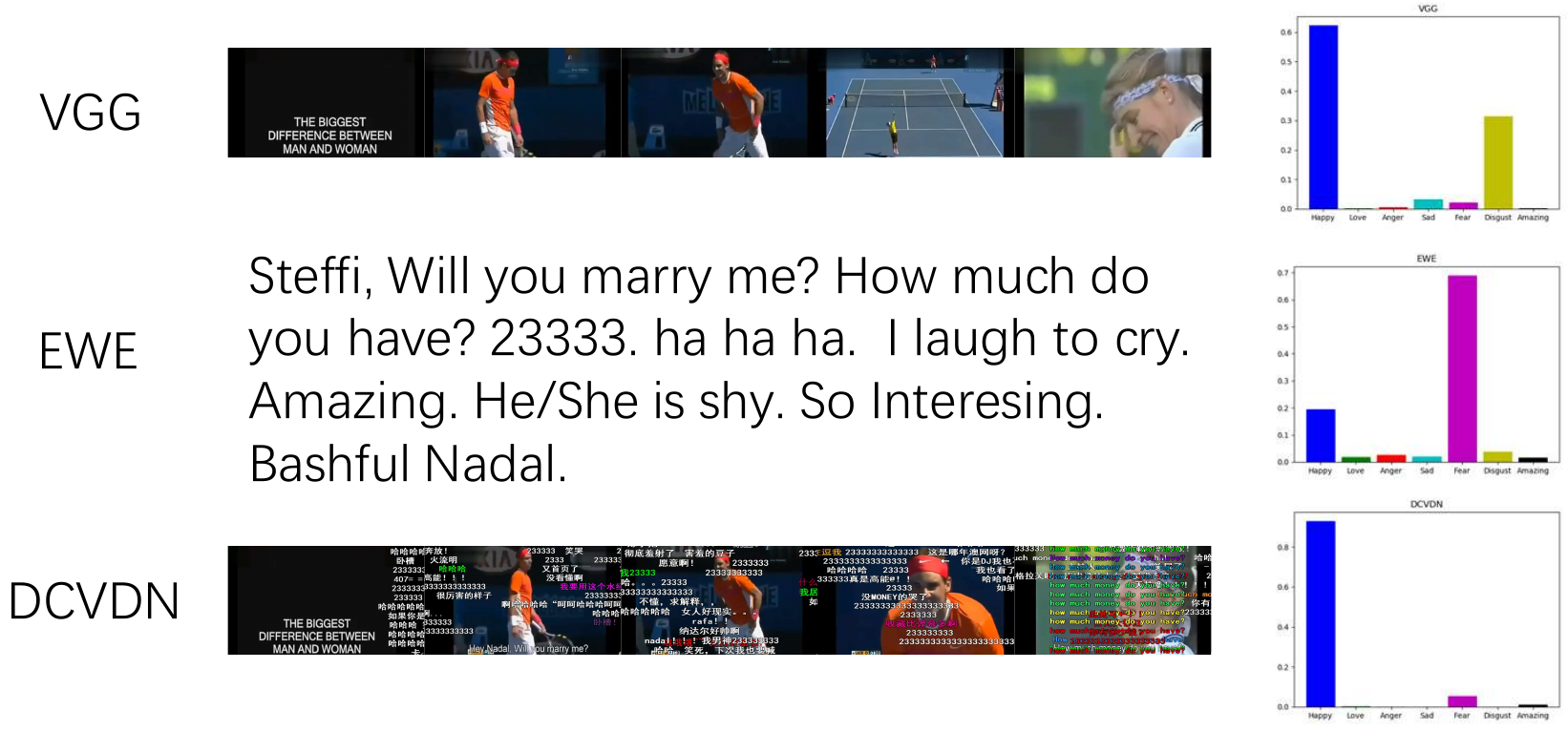}}
            \subfigure[A combination of several movie clips with sad BGM. The label is ``Sad''. EWE and DCVDN predict correctly, but VGG considers it's ``Happy''.]{\label{fig:subfig:c}
                \includegraphics[width=0.9\textwidth]{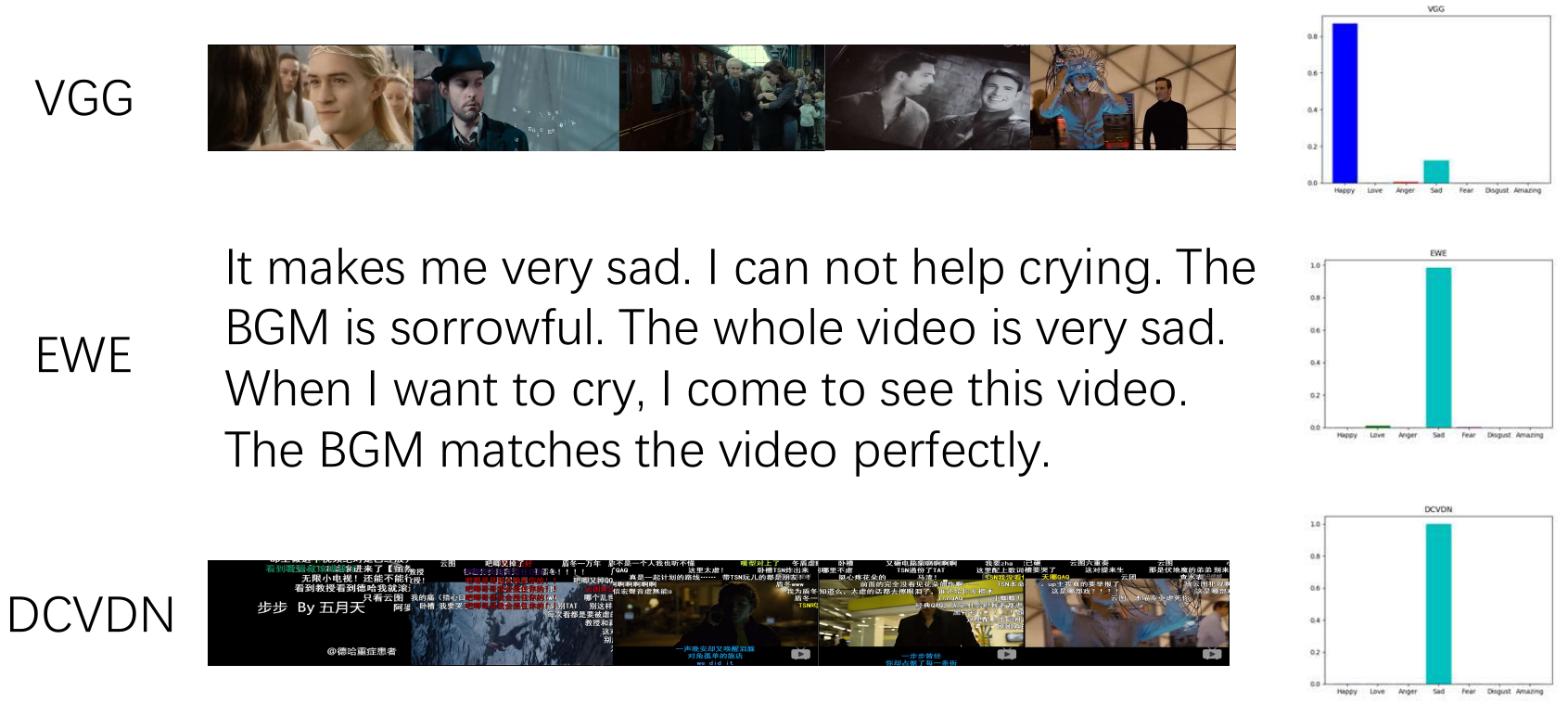}}
            \caption{Three prediction examples to illustrate performance comparison between VGG, EWE and DCVDN. }
            \label{fig:example_prediction}
        \end{figure}

        In the subsubsection, we provide three prediction examples to illustrate the performance comparison between VGG (visual method only), EWE (textual method only) and DCVDN (our method, jointly combining visual and textual information), as shown in Fig. \ref{fig:example_prediction}.

        Fig. \ref{fig:example_prediction} (a) is a wacky music video with common visual content, however, the background music (BGM) sounds scaring.
        The ground truth of this video is ``Disgust''. DCVDN gives the right answer. VGG considers it's ``Love'' for it looks like a music video and most music video make us feel Love. The result of EWE is ``Fear'', the reason of which is that the audiences sometimes express their disgust via the adjective of fear in texts, such as "Help me! I died as soon as he sang. I choose to die".
        Fig. \ref{fig:example_prediction} (b) is a video about two tennis stars, Nadal and Steffi. At the beginning of the video, one audience asked Nadal, "Will you marry me?".
        Nadal refused her shyly.
        Then, another audience asked Steffi, "Will you marry me?". Steffi asked back, "How much do you have?". Other audiences in the stands laughed loudly.
        The ground truth of example (b) is ``Happy'', as the audiences propose to the stars are very funny.
        VGG and DCVDN all give the right answer, probably because the video is about sports stars, while the probability of Disgust ranks second high in the result of VGG maybe due to the poor quality of videos.
        EWE considers the emotion of the video ``Fear''. This probably because word ``shy'' appears frequently in danmu texts and ``Shy'' is the subclass of ``Fear'' in our dataset.
        Fig. \ref{fig:example_prediction} (c) is a music video combination clips from several movies, such as Harry Potter and the Lord of the Rings.
        These clips with sorrowful BGM are more about the parting or about death, and the ground truth is ``Sad''.
        Both of EWE and DCVDN give the right answer with almost $100\%$ probability, while VGG considers it's ``Happy'' with high probability and it's ``Sad'' with low probability.
        These results are reasonable because the visual content is mainly about the movie stars, which looks ``Happy'' in most time.
        However, in these cases, danmu would give us more information about the true feeling of audiences, which is beyond the visual content.

\subsection{Evaluations}
    \subsubsection{EWE on the Emotional Analysis}
    We first compare our EWE model with textual-based baselines on our own dataset and two public datasets.
    Table \ref{tab:text} shows the comparison results with the texts in our own dataset. EWE outperforms all other textual-based baselines under investigation by $2.09\%$ to $219.16\%$ on \textit{Accuracy}. The performance of the lexicon-based method and eLDA is pretty poor, which indicates that the relation between the number of emotional words and the emotion label of videos are not that strong. The embedding-based methods can perform much better, which can effectively capture the upper-level features in emotion space through highly non-linear transformations. EWE achieves the best performance, which is due to the fact that the emotional word embeddings are more informative and could provide more hints for emotion analysis.
    Table \ref{tab:text:isear} shows the comparison results on the ISEAR dataset. EWE performs more steadily than all other textual-based baselines with $10.92\%$ to $167.57\%$ improvement on \textit{Accuracy}. EWE may perform even better if the training examples in the dataset is with more balanced distribution across different classes.
    Table \ref{tab:text:multi-domains} demonstrates the comparison results on the Multi-Domain Sentiment dataset, which only contains the positive and negative labels. Besides the baselines investigated with previous two datasets, we also include SSWE \cite{tang2014learning} as the baseline for the sentiment classification task. The performance of EWE is also the best one, with $1.40\%$ to $28.91\%$ improvement on \textit{Accuracy}, although it's relatively not that outstanding like the performance on previous two datasets. We notice that the performance of lexicon method and eLDA are bad, which may indicate that the quality of the sentiment dictionary is not good. This could adversely dampen the performance of EWE, which could be the possible reason for EWE not prominent as with the previous two datasets.

    \subsubsection{DCVDN-V on the Video-Danmu Dataset}


		\begin{table}
    			\centering
    			\caption{Comparison of \textit{Precision} and \textit{Accuracy} between EWE and textual baselines on the Video-Damu dataset.}
    			\begin{tabular}{|c|c|c|c|c|c|}
    				\hline
    				\textit{Precision} & Lexcion & eLDA & WE & TWE & EWE\\
    				\hline
    				happy & 0.106 & 0.493 & 0.568 & 0.644 & 0.636 \\
    				\hline
    				love & 0.757 & 0.051 & 0.737 & 0.749 & 0.777 \\
    				\hline
    				anger & 0.0 & 0.0 & 1.0 & 1.0 & 1.0 \\
    				\hline
    				sad & 0.067 & 0.547 & 0.803 & 0.826 & 0.811\\
    				\hline
    				fear & 0.049 & 0.223 & 0.384 & 0.624 & 0.504\\
    				\hline
    				disgust & 0.094 & 0.659 & 0.554 & 0.647 & 0.630\\
    				\hline
    				surprise & 0.12 & 0.062 & 0.299 & 0.688 & 0.403\\
    				\hline\hline
    				\textit{\textbf{Accuracy}} & 0.214 & 0.321 & 0.624 & 0.669 & \textbf{0.683} \\
    				\hline
  			 \end{tabular}
    			\label{tab:text}
		\end{table}

        \begin{table}
                \centering
                \caption{Comparison of \textit{Precision} and \textit{Accuracy} between EWE and textual baselines on ISEAR.}
                \begin{tabular}{|c|c|c|c|c|c|}

                    \hline
                    \textit{Precision} & Lexcion & eLDA      & WE    & TWE   & EWE\\
                    \hline
                    Anger              & 0.131   & 0.274    & 0.258 & 0.234 & 0.309 \\
                    \hline
                    Disgust            & 0.139   & 0.308    & 0.339 & 0.313 & 0.346 \\
                    \hline
                    Joy                & 0.154     & 0.186      & 0.428 & 0.451 & 0.472 \\
                    \hline
                    Shame              & 0.148   & 0.165    & 0.244 & 0.309 & 0.244\\
                    \hline
                    Fear               & 0.150   & 0.445    & 0.414 & 0.391 & 0.244\\
                    \hline
                    Sadness            & 0.161   & 0.359    & 0.437 & 0.515 & 0.488\\
                    \hline
                    Guilt              & 0.152    & 0.0    & 0.312 & 0.304 & 0.488\\
                    \hline\hline
                    \textit{\textbf{Accuracy}} & 0.148 & 0.272 & 0.354 & 0.357 & \textbf{0.396} \\
                    \hline
             \end{tabular}
                \label{tab:text:isear}
        \end{table}

        \begin{table}
                \centering
                \caption{Comparison of \textit{Precision} and \textit{Accuracy} between EWE and textual baselines on Multi-Domain Sentiment Dataset.}
                \begin{tabular}{|c|c|c|c|}
                    \hline
                    \textit{Precision} & positive & negative & \textit{Accuracy}\\
                    \hline
                    Lexcion            & 0.510    & 0.516    & 0.512 \\
                    \hline
                    eLDA                & 0.505    & 0.505    & 0.505 \\
                    \hline
                    SSWE               & 0.560    & 0.624    & 0.613 \\
                    \hline
                    WE                 & 0.569    & 0.680    & 0.639 \\
                    \hline
                    TWE                & 0.560    & 0.678    & 0.642 \\
                    \hline
                    EWE                & 0.580    & 0.680    & 0.651  \\
                    \hline
             \end{tabular}
                \label{tab:text:multi-domains}
        \end{table}
        We compare the visual part of our model, DCVDN-V, with other visual-based baselines and video classification methods on the Video-Danmu dataset. DCVDN-V is the reduced version of DCVDN solely considering visual input and using VGG-Net and autoencoder for feature extraction.
        Table \ref{tab:visual} shows the comparison results between visual-based baselines and DCVDN-V. Similarly, the \textit{Precision} is counted based on each respective emotion class and the \textit{Accuracy} is the overall average across all emotion classes.
        Basically, DCVDN-V outperforms other visual-based baselines by $16.92\%$ to $35.03\%$ on \textit{Accuracy}. Moreover, the deep learning based methods generally achieve more or less improvements compared with the low-level feature based approach. It is also worth pointing out that the \textit{Precision} of ``Happy'' and ``Love'' predicted by visual-based methods is relatively lower than other classes compared with textual-based methods.
        The reason may be that the visual characteristics of ``Happy'' and ``Love'' videos are quite similar to each other so that the features may lead to great intra-class variance. This phenomenon strongly verify that it is hard to learn a clear mapping function solely from visual features to high-level emotions. Therefore, with the enhancement by the interactive information from user-generated texts, joint features may achieve remarkable improvements compared with pure visual-based methods. As our dataset is based on videos, we also compare DCVDN-V with other video classification baselines, with the result shown in Table \ref{tab:video-classification}. The results show that the performance of different video classification methods are very close to each other, while DCVDN-V outperforms them significantly by $22.03\%$ to $25.77\%$ enhancement on \emph{Accuracy}. This demonstrates that our method can learn more information related to emotion analysis rather than the video classification methods.
        What's more, the outperformance of DCVDN-V in these experiments can show that the features if the videos are learned well with the help of the mutual information from the text part.


		   \begin{table}
                \centering
                \caption{Comparison of \textit{Precision} and \textit{Accuracy} between DCVDN-V and visual baselines.}
                \begin{tabular}{|c|c|c|c|c|c|c|c|c|}
                    \hline
                    \textit{Precision} & GCH   & LCH   & PCNN  & CaffeNet & DCVDN-V \\
                    \hline
                    happy              & 0.283 & 0.170 & 0.271 & 0.174    & 0.323 \\
                    \hline
                    love               & 0.361 & 0.355 & 0.355 & 0.405    & 0.463 \\
                    \hline
                    anger & 0.833 & 0.573 & 0.938 & 0.963 & 1.0 \\
                    \hline
                    sad & 0.364 & 0.384 & 0.440 & 0.541 & 0.616 \\
                    \hline
                    fear & 0.359 & 0.438 & 0.438 & 0.481 & 0.609 \\
                    \hline
                    disgust & 0.448 & 0.452 & 0.411 & 0.518 & 0.642 \\
                    \hline
                    surprise & 0.219 & 0.343 & 0.346 & 0.272 & 0.338 \\
                    \hline\hline
                    \textit{\textbf{Accuracy}} & 0.410 & 0.394 & 0.423 & 0.455 & \textbf{0.532} \\
                    \hline
            \end{tabular}
                \label{tab:visual}
        \end{table}

        \begin{table}
    			\centering
    			\caption{Comparison of \textit{Precision} and \textit{Accuracy} between DCVDN-V and video classification baselines.}
    			\begin{tabular}{|c|c|c|c|c|c|c|c|c|}
    				\hline
    				\textit{Precision} & Conv3D & Temporal & T + S & DCVDN-V \\
    				\hline
    				happy              & 0.275  & 0.313    & 0.308              & 0.323 \\
    				\hline
    				love               & 0.366  & 0.446    & 0.338              & 0.463 \\
    				\hline
    				anger              & 0.941  & 0.940    & 0.968              & 1.0 \\
    				\hline
    				sad                & 0.472  & 0.489    & 0.667              & 0.616 \\
    				\hline
    				fear               & 0.430  & 0.407    & 0.387              & 0.609 \\
    				\hline
    				disgust            & 0.417  & 0.433    & 0.524              & 0.642 \\
    				\hline
    				surprise           & 0.439  & 0.345    & 0.232              & 0.338 \\
    				\hline\hline
    				\textit{\textbf{Accuracy}} & 0.436 & 0.423 & 0.427             & \textbf{0.532} \\
    				\hline
   			\end{tabular}
    			\label{tab:video-classification}
		\end{table}

        \subsubsection{DCVDN on the Video-Danmu Dataset}

        Table \ref{tab:multi-view} shows the comparison results between DCVDN and multi-view learning baselines, where the `` T + S" means ``Temporal + Spatial" in the first line of the fourth column.
        The proposed DCVDN with DCCAE surpasses other multi-view learning methods by $1.53\%$ to $2.52\%$ on \textit{Accuracy}. The performance of DistAE sometimes is even worse than Simple-Con. This is because DistAE aims to minimize the distance between visual and textual views, however they are not the same although they are somewhat correlated. By contrast, DCCAE provides the flexibility to dig deep about the relationship between different views so as to effectively facilitate joint representation learning. The whole results shown in Table \ref{tab:multi-view} can also justify that CCA is able to maximize the mutual information between videos and danmus.

		\begin{table}
		\small
    			\centering
    			\caption{Comparison of \textit{Precision} and \textit{Accuracy} between DCVDN and multi-view learning baselines.}
    			\begin{tabular}{|c|c|c|c|}
    				\hline
    				\textit{Precision} & Simple-Con & DistAE & DCVDN  \\
    				\hline
    				happy  & 0.729 & 0.622 & 0.732\\
    				\hline
    				love & 0.816 & 0.782 & 0.754 \\
    				\hline
    				anger & 1.0 & 1.0 & 1.0 \\
    				\hline
    				sad & 0.795 & 0.805 & 0.814 \\
    				\hline
    				fear & 0.632 & 0.571 & 0.716 \\
    				\hline
    				disgust & 0.601 & 0.627 & 0.628 \\
    				\hline
    				surprise & 0.442 & 0.652 & 0.450\\
    				\hline\hline
    				\textit{\textbf{Accuracy}} & 0.720 & 0.713 & \textbf{0.731}\\
    				\hline
    			\end{tabular}
    			\label{tab:multi-view}
		\end{table}

        \subsubsection{Impact of the Size of Dataset}
        In this subsection, we show that the size of our dataset is large enough to learn a good model. We test the accuracy with on different ratios of the size of our dataset. The accuracy is $67.8\%$, $71.3\%$, $72.7\%$ respectively when the ratio of size is $0.4$, $0.6$ and $0.8$, which can show that our dataset is large enough to prove the performance of our models.

\section{Conclusions}
	In this paper, we study user emotion analysis toward online videos by jointly utilizing video frames and danmu texts simultaneously.
        	To encode emotion into the learned word embeddings, we propose EWE to learn text representations by jointly considering their semantics and emotions.
    	Afterwards, we propose a novel visual-textual emotion analysis approach with deep coupled video and danmu neural networks, in which visual and textual features are synchronously extracted and fused to form a comprehensive representation by deep-canonically-correlated-autoencoder-based multi-view learning.
    	To evaluate the performance of EWE and DCVDN, we conduct extensive experiments on public datasets and self-crawled video-damu dataset.
    	The experimental results strongly validate the superiority of EWE and the overall DCVDN over other state-of-the-art baselines.

\bibliographystyle{plain}
\bibliography{reference}

\end{document}